\documentclass[11pt,a4paper]{article}
\usepackage[hyperref]{emnlp2018}
\usepackage{times,graphicx}
\usepackage{latexsym,booktabs,multicol,multirow,colortbl}
\usepackage{url}
\usepackage[]{todonotes}   % insert [disable] to disable all notes

\aclfinalcopy % Uncomment this line for the final submission
\title{Why is unsupervised alignment of English embeddings from different algorithms so hard?}

\author{Mareike Hartmann \\
  Dep. of Computer Science \\
  University of Copenhagen \\
  Denmark \\
  {\tt hartmann@di.ku.dk} \\\And
  Yova Kementchedjhieva \\
  Dep. of Computer Science \\
  University of Copenhagen \\
  Denmark \\
  {\tt yova@di.ku.dk} \\
  \\\And
   Anders S{\o}gaard  \\
  Dep. of Computer Science \\
  University of Copenhagen \\
  Denmark \\
  {\tt soegaard@di.ku.dk} \\}

\date{}

\begin{document}
\maketitle
\begin{abstract}
This paper presents a challenge to the community: Generative adversarial networks (GANs) can perfectly align independent English word embeddings induced using {\em the same}~algorithm, based on distributional information alone; but fails to do so, for two different embeddings algorithms. {\em Why is that?} We believe understanding why, is key to understand {\em both} modern word embedding algorithms {\em and } the limitations and instability dynamics of GANs. This paper shows that (a) in all these cases, where alignment fails, there exists a linear transform between the two embeddings (so algorithm biases do not lead to non-linear differences), and (b) similar effects can not easily be obtained by varying hyper-parameters. One plausible suggestion based on our initial experiments is that the differences in the inductive biases of the embedding  algorithms lead to an optimization landscape that is riddled with local optima, leading to a very small basin of convergence, but we present this more as a challenge paper than a technical contribution.

%We study the limitations of unsupervised bilingual dictionary algorithms -- learning to align monolingual embeddings by exploiting common, global structure -- by starting with the seemingly trivial case of unsupervised alignment of two embedding spaces induced from identical corpora. We observe, however, that if the algorithms are learned by different embedding algorithms, unsupervised alignment fails, suggesting that embedding algorithms do not learn unbiased representations of global structure. We then manipulate the target corpus and/or the target embeddings in several ways: scrambling sentences, normalizing vectors, changing embedding algorithm, subsampling the data, etc. These controlled experiments on semi-synthetic enable us to study the limitations of unsupervised bilingual dictionary algorithms, finding that XXX is key to their performance.  
\end{abstract}

\section{Introduction}

This paper brings together two fascinating research topics in natural language processing (NLP), namely {\em understanding the properties of word embeddings} \cite{Mikolov2013,Mitchell:Steedman:15,Mimno:Thompson:17} and {\em unsupervised bilingual dictionary induction} \cite{Conneau:ea:18,Zhang2017,Soegaard2018}. In an effort to better understand when unsupervised bilingual dictionary induction is possible, we factored out linguistic differences between languages, and studied English-English alignability (by learning to align English embeddings trained on different samples of the English Wikipedia), when we came across a puzzling phenomena: {\em English-English can be aligned with almost 100\%~precision, if you use the same embedding algorithms for the two samples, but not at all (0\%~precision), if you use different embedding algorithms}. This results suggest that the properties of word embeddings induced by different algorithms challenge unsupervised bilingual dictionary algorithms. Understanding why will enable us to develop more stable adversarial learning algorithms and give us a better understanding of how embedding algorithms differ. 

\paragraph{Contributions} We are, to the best of our knowledge, the first to study unsupervised alignability of pairs of English word embeddings. We show that unsupervised alignment -- specifically the MUSE system \cite{Conneau:ea:18} -- fails when the algorithms used to induce the two embeddings differ, and that this is {\em not} because there is no linear transformation between the two spaces. We further show that poor initialization, as a result of MUSE initially applying an identity transform to two word embeddings far apart in space, is not the sole reason the discriminator suffers from local optima. Finally, we present an experiment showing what the minimal corpus size is for unsupervised alignment to succeed, in the absence of linguistic differences. 

\section{Aligning embeddings}

\subsection{Unsupervised alignment using generative adversarial networks}

MUSE \cite{Conneau:ea:18} uses a vanilla generative adversarial network (GAN) with a linear generator to learn alignments between embedding spaces without supervision. In a two-player game, a discriminator $D$ aims to tell the two language spaces apart, while a generator $G$ aims to map the source language into the target language space, fooling the discriminator. While MUSE achieves impressive results at times, MUSE is highly unstable, e.g., with different initializations precision scores vary between 0\% and 45\% for English-Greek \cite{Soegaard2018}.

The parameters of a GAN with a linear generator are $(\Omega, w)$. They are obtained by solving the following min-max problem:

\begin{equation}\label{minmax}\min_\Omega\max_w
\mathrm{E}[\log(D_w(X)) + \log(1 - D_w(g_\Omega(Z)))]\end{equation} which reduces to \begin{equation}\label{jsd}\min_\Omega\mathit{JS}(P_X\mid P_\Omega)\end{equation}

$\Omega$ is initialized as the identity matrix $I$. 

If $G$ wins the game against an ideal discriminator on a very large number of samples, then $F$ (the source vector space) and $\Omega E$ (with $E$ being the target vector space) can be shown to be close in Jensen-Shannon divergence, and thus the model has learned the true distribution. This result, referring to the distributions of the data, $p_{\mathit{data}}$, and the distribution, $p_g$, $G$ is sampling from, is from \citet{Goodfellow2014}: {\em If $G$ and $D$ have enough capacity, and at each step of training, the discriminator
is allowed to reach its optimum given $G$, and $p_g$ is updated so as to improve the criterion
$$E_{\mathbf{x}\sim p_{\mathit{data}}}[
	\log D_G^{\ast}(\mathbf{x})] 
+ E_{\mathbf{x}\sim p_g} [\log(1 - D^{\ast}_G(\mathbf{x}))]$$
then $p_g$ converges to $p_{\mathit{data}}$.}

This result relies on a number of assumptions that do not hold in practice. Our generator, which learns a linear transform $\Omega$, has very limited capacity, for example, and we are updating $\Omega$ rather than $p_g$. In practice, therefore, during training, we alternate between $k$ steps of optimizing
the discriminator and one step of optimizing the generator. If the GAN-based alignment is not successful, this can thus be a result of two things: Either that $G$ does not have enough capacity, or that $D$ is stuck in a local optimum. Our results in \S3 show that the inability to align English-English in the case of different word embedding algorithms is {\em not}~a result of limited capacity, but a result of the GAN being trapped in %what looks like a bit of a mine field of local optima.
one of the many local optima of the loss function.

%These points suggests that it is often better to rely on a seed dictionary---if available---or a dictionary based on identical strings to initialize Procrustes instead of one based on adversarial training. Given the convincing performance of (supervised) Procrustes based on identical string alignments \cite{Søgaard2018}, we will focus on improving Procrustes in the following\footnote{Our findings should also apply to Procrustes using a dictionary based on adversarial training}. 

%The above observations are not aimed to discredit adversarial training as an appropriate method in solving the task of BDI.  In lack of any seed data, any orthographic overlap between the languages involved and, in the extreme case, of any overlap in numerals, adversarial training really seems to be the only viable option.
%Yet, it is worth noting that for a vast number of languages these conditions don't necessarily hold. Seeing that Procrustes alone, given a seed of gold-standard translational pairs, can achieve as much or more than a two-step system involving adversarial training, for the rest of this paper we shall focus solely on Procrustes.
% Notice that any findings relating to Procrustes in a one-step system, should also hold for it as refinement procedure in a two-step system. [Could also just keep the very last sentence and scrape everything else.] 

\subsection{Supervised alignment using Procrustes Analysis}
Procrustes Analysis \cite{Schonemann:66} has been commonly used for supervised alignment of word embeddings \cite{Smith2017, Artetxe2018}. Here, the optimal alignment between two embedding spaces is computed using singular value decomposition of the aligned embeddings in a seed dictionary. \citet{Conneau:ea:18} use Procrustes Analysis to refine an initial seed dictionary learned by the generative adversarial network without supervision. 
In our supervised experiments, we use 5000 seed words as supervision for learning the alignment between embeddings.
\subsection{Geometry of embeddings}

Below we summarize some previous findings about the geometry of monolingual embeddings \cite{Mimno:Thompson:17}, and add some new observations. We discuss five embedding algorithms: SVD on positive PMI matrices (Hyperwords-SVD) \cite{Levy15}, skip-gram negative sampling applied to co-occurrence matrices (Hyperwords-SGNS) \cite{Levy15}, continuous bag-of-words (CBOW) \cite{Mikolov2013a}, GloVe \cite{Pennington14}, and FastText \cite{bojanowski2017}. To analyze the geometry of our monolingual embeddings in space, we report average inner product to mean vector; see \newcite{Mimno:Thompson:17} for details. 

\paragraph{Hyperwords-SVD} have a small average inner product (0.0032), suggesting they are well-dispersed through space; like Hyperwords-SGNS and standard SGNS \cite{Mimno:Thompson:17}, they do not exhibit a clear word frequency bias. \textbf{Hyperwords-SGNS} vectors also have a small average inner product (0.0002), in contrast with standard SGNS vectors, which are narrowly clustered
in a single orthant \cite{Mimno:Thompson:17}. In line with standard SGNS vectors, the frequency 
of words has relatively little effect on their
inner product, with the exception of the rare words,
which have slightly less positive inner products. \textbf{CBOW} vectors have a relatively large average inner product (4.2985). The vectors trained by \textbf{GloVe} show a
clear relationship with word frequency, with low-frequency
words opposing the frequency-balanced
mean vector. The embeddings are well-dispersed, with an average inner product of 0.0002. Finally, \textbf{FastText} vectors
have a large, positive inner product with the
mean (0.2988), indicating that they are not evenly dispersed
through the space, but pointing in roughly the same direction. The FastText vectors exhibit a frequency bias, much like what has been previously observed with GloVe vectors. The differences are the results of the inductive biases of the different embedding algorithms.

\begin{table*}
\begin{center}\begin{small}
\begin{tabular}{l|r|r|r|r|r}%|ll|ll}
\toprule
&\multicolumn{1}{c}{Hyperwords-SGNS}&\multicolumn{1}{c}{Hyperwords-SVD}&\multicolumn{1}{c}{CBOW}&\multicolumn{1}{c}{GloVe}&\multicolumn{1}{c}{FastText}\\
\midrule
\multicolumn{6}{c}{\sc Unsupervised}\\
\midrule
\multirow{1}{*}{Hyperwords-SGNS}&{\bf 0.997}&\cellcolor{gray!25}&\cellcolor{gray!25}&\cellcolor{gray!25}&\cellcolor{gray!25}\\% &&&&\\
%&&&&\\% &&&&\\
\midrule
\multirow{1}{*}{Hyperwords-SVD}&0.000&{\bf 0.992}&\cellcolor{gray!25}&\cellcolor{gray!25}&\cellcolor{gray!25}\\% &&&&\\
%&&&&\\% &&&&\\
\midrule
\multirow{1}{*}{CBOW}&0.000&0.000&{\bf 0.997}&\cellcolor{gray!25}&\cellcolor{gray!25}\\% &&&&\\
%&&&&\\% &&&&\\
\midrule
\multirow{1}{*}{GloVe}&0.000&0.000&0.000&{\bf 0.997}&\cellcolor{gray!25}\\% &&&&\\
%&&&&\\% &&&&\\
\midrule
\multirow{1}{*}{FastText}&0.000&0.000&0.000&0.000&{\bf 0.997}\\% &&&&\\
%&&&&\\% &&&&\\
\midrule\midrule
\multicolumn{6}{c}{\sc Supervised}\\
%\midrule
%&\multicolumn{1}{c}{Hyperwords-sgns}&\multicolumn{1}{c}{Hyperwords-SVD}&\multicolumn{1}{c}{CBOW}&\multicolumn{1}{c}{GloVe}&\multicolumn{1}{c}{FastText}\\
\midrule
%\multirow{1}{*}{Hyperwords-sgns}&{\bf 0.997}&0.000&0.000&0.000&0.000\\% &&&&\\
%&&&&\\% &&&&\\
%\midrule
\multirow{1}{*}{Hyperwords-SVD}&0.967&\cellcolor{gray!25}&\cellcolor{gray!25}&\cellcolor{gray!25}&\cellcolor{gray!25}\\% &&&&\\
%&&&&\\% &&&&\\
\midrule
\multirow{1}{*}{CBOW}&0.990&0.989&\cellcolor{gray!25}&\cellcolor{gray!25}&\cellcolor{gray!25}\\% &&&&\\
%&&&&\\% &&&&\\
\midrule
\multirow{1}{*}{GloVe}&0.985&0.992&0.999&\cellcolor{gray!25}&\cellcolor{gray!25}\\% &&&&\\
%&&&&\\% &&&&\\
\midrule
\multirow{1}{*}{FastText}&0.994&0.994&0.999&0.997&\cellcolor{gray!25}\\% &&&&\\
%&&&&\\% &&&&\\
\bottomrule
\end{tabular}\end{small}
\caption{Precision at 1 (P@1) for unsupervised GAN alignment with Procrustes refinement (top) and supervised Procrustes analysis for the cases in which unsupervised alignment fails (bottom). Results clearly show that GANs can align two independent embeddings induced by the same algorithm; but not embeddings aligned by different ones. Supervised Procrustes analysis, on the other hand, perfectly aligns the embeddings in both cases. }\label{t:algvsalg}
\end{center}
\end{table*}

\section{Experiments}

This section presents our data, the hyper-parameters of our embeddings, our experimental protocols, and our results. 

\subsection{Data}
In the following experiments we learn word embeddings on samples of a publicly available Wikipedia dump from March 2018.\footnote{\url{https://dumps.wikimedia.org/enwiki/}} 
The data is preprocessed using a publicly available preprocessing script\footnote{\url{http://mattmahoney.net/dc/textdata.html}}, extracting text,  removing non-alphanumeric characters, converting digits to text, and lowercasing the text.

\subsection{Hyper-parameters}
We train 300-dimensional word embeddings using the algorithms' recommended hyperparameter settings, listed in the following:\footnote{We also ran experiments with one of the embedding algorithms (FastText) to check if our results were robust across hyper-parameter settings}~ For \textbf{Hyperwords-SGNS}, the window size is set to 2 and the subsampling of frequent words and smoothing of the context distribution are disabled. The minimal word count for being in the vocabulary is 100. The same applies for \textbf{Hyperwords-SVD}, and the exponent for weighting the eigenvalue matrix is 0.5. For \textbf{CBOW}, the window size is set to 8, the number of negative samples is 25, and the subsampling threshold for frequent words  is 1e-4. For \textbf{GloVe}, the window size is set to 15 and the cutoff parameter x$_{max}$ to 10.  
Finally, for \textbf{FastText}, the window size is 5, the number of negatives samples is 5 and the sampling threshold is 0.0001.

\subsection{Main experiments}

We train word embeddings using the different embedding algorithms listed in \S3.2 on two non-overlapping 10\% samples of the English Wikipedia dump (the samples contain 463,576 and 528,556 distinct words, with an overlap in vocabulary of 351,858 words). We learn unsupervised and supervised alignments for embeddings (as described in \S2) trained by different algorithms on the same datasplits, and for embeddings trained by the same algorithm on the two different datasplits. For the unsupervised alignments, we use the default parameters of the MUSE system for the adversarial training, i.e. a discriminator with 2 fully connected layers of 2048 units trained over 5 epochs, 1,000,000 iterations per epoch with 5 discriminator steps per iteration and a batch size of 32. 

We evaluate the alignments in terms of Precision@1 in the word translation retrieval task for the 1500 test words used by \citet{bojanowski2017}. The results are shown in Table \ref{t:algvsalg}\footnote{We report Precision at 1 scores but find that the pattern is the same for Precision at 10, with perfect alignments for embeddings from the same algorithm and 0 scores for alignments between embeddings from different algorithms in the unsupervised experiments.}. Our main observations are: (a) MUSE learns perfect alignments for embeddings learned by the same algorithm on different data splits. (b) MUSE cannot learn alignments for embeddings learned by different algorithms on the same data splits, even if there exists a linear transformation aligning both sets of embeddings (the supervised algorithm learns perfect alignments). We also verify that MUSE cannot learn to align embeddings from different algorithms {\em even when induced from the same sample}. As already mentioned, we also ran experiments to check that the failure of MUSE to learn good alignments was not a result of the differences in hyper-parameter settings. \S3.4 presents additional experiments with normalization, for control; \S3.5 addresses how much data is needed to align independently induced embeddings from the same algorithm. \S4 discusses potential answers to why MUSE fails when embeddings are induced using different algorithms.

\subsection{Experiments with normalization}

The embeddings in the main experiments differ in several ways; see \S2. One possible explanation for the inability of MUSE to align embeddings from different algorithms could be that the two embeddings are so far apart in space that the discriminator learns to discriminate between them too quickly. Recall that $\Omega$ is initialized as the identity matrix $I$, which means that the generator initially presents the discriminator with the source embedding as is. This is an effect that has often been observed with GANs \cite{Arjovsky:Bottou:17}; could this also be the explanation for our results? At a first glance, this seems a possible explanation. The inner products with the mean differ significantly for the five embedding algorithms (see \S2). The only embeddings that have roughly the same directionality are Hyperwords and GloVe, and their centroids are very far apart in cosine space. The cosine similarity of the centroids of the two versions of Hyperwords is -0.006, and the cosine similarity for Hyperwords-SVD and GloVe is 0.019. 
%\begin{table}\begin{center}\begin{tabular}{ll}
%\toprule
%Hyperwords-SGNS&0.0032\\
%Hyperwords-SVD&0.0002\\
%CBOW&4.2985\\
%GloVe&0.0002\\
%FastText&0.2988\\
%\bottomrule
%\end{tabular}
%\caption{Spread of embeddings (averages of two 10\%~samples).}
%\end{center}\end{table}
%ft_1	0.286117843
%ft_2	0.3113868692
%cbow_1	4.187370679
%%cbow_2	4.39196044
%glove_1	0.3881540808
%glove_2	0.308782316
%svd_1	0.0001755654923
%svd_2	0.0001959642567
%sgns_1	0.003401878739
%sgns_2	0.003068093511
However, poor initialization as a result of applying the identity transform to very distant word embeddings is not the explanation for the poor performance of MUSE in this set-up: Both sets of Hyperwords embeddings were normalized, but alignment still failed. To verify this holds in general, i.e., that results are not affected by normalization in general, we also ran experiments with the remaining 14 embedding pairs, normalizing and/or centering both embeddings. Results stayed the same: Precision at 1 scores of 0.

\begin{figure}
\includegraphics[width=1\linewidth]{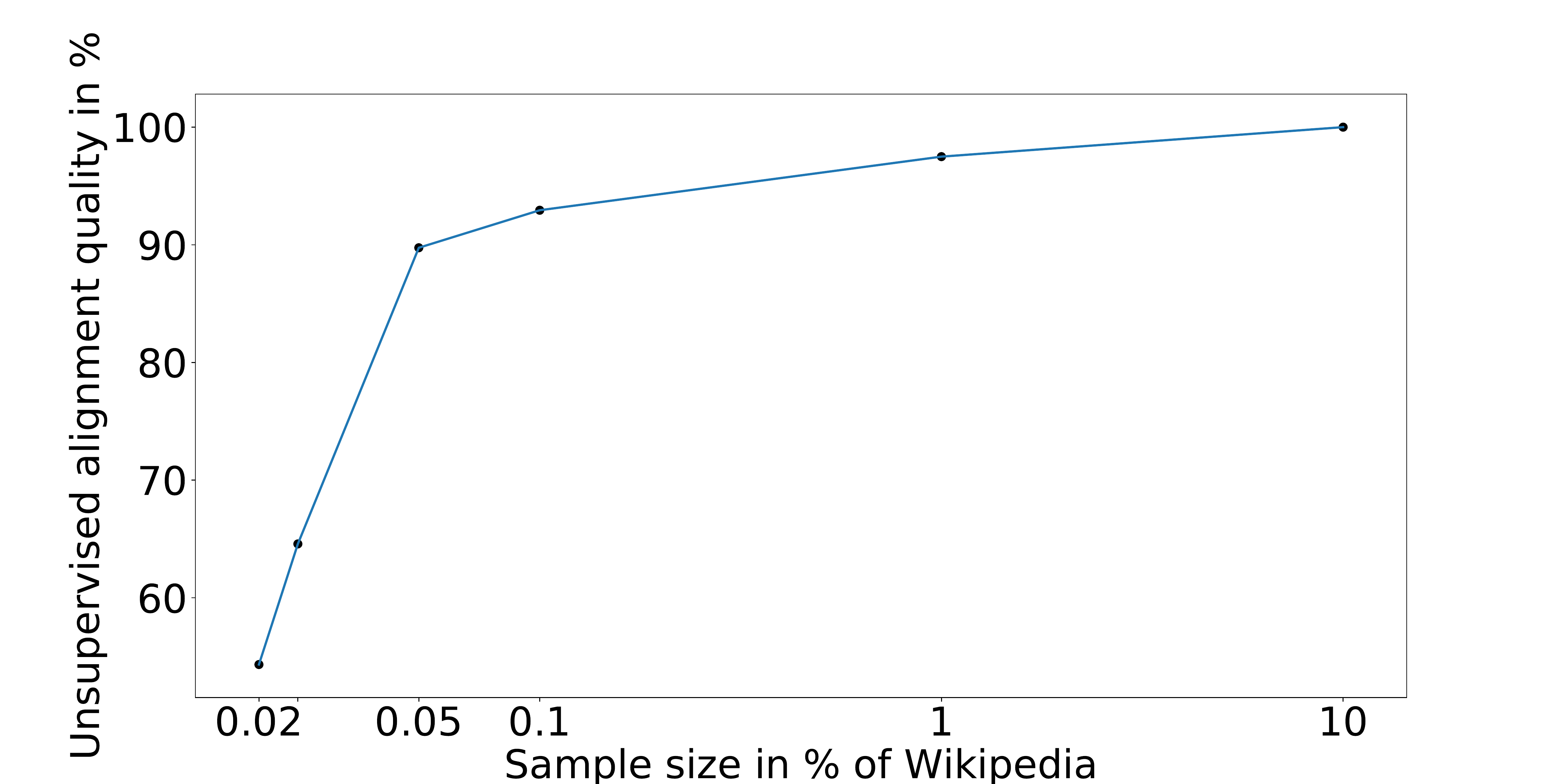}
\caption{Unsupervised alignment quality for FastText embeddings trained on samples of different sizes, evaluated on 878 words covered by all of the embeddings. The x-axis is log-scaled.}\label{f:learningcurve}
\end{figure}

\subsection{Learning curve}

MUSE perfectly aligns independently induced word embeddings induced by the same algorithm. For FastText, it correctly aligns 99.7\%~of all words in the evaluation lexicon with itself. Our samples are 10\%~of a publicly available Wikipedia dump, amounting to more than 400 million tokens per sample. English-English alignment is an interesting control experiment for unsupervised bilingual dictionary induction, abstracting away from linguistic differences, and we ran a series of experiments to see how small samples MUSE can align in the absence of linguistic differences. The learning curve is presented in Figure~\ref{f:learningcurve}. 

\begin{figure}
\includegraphics[width=1\linewidth]{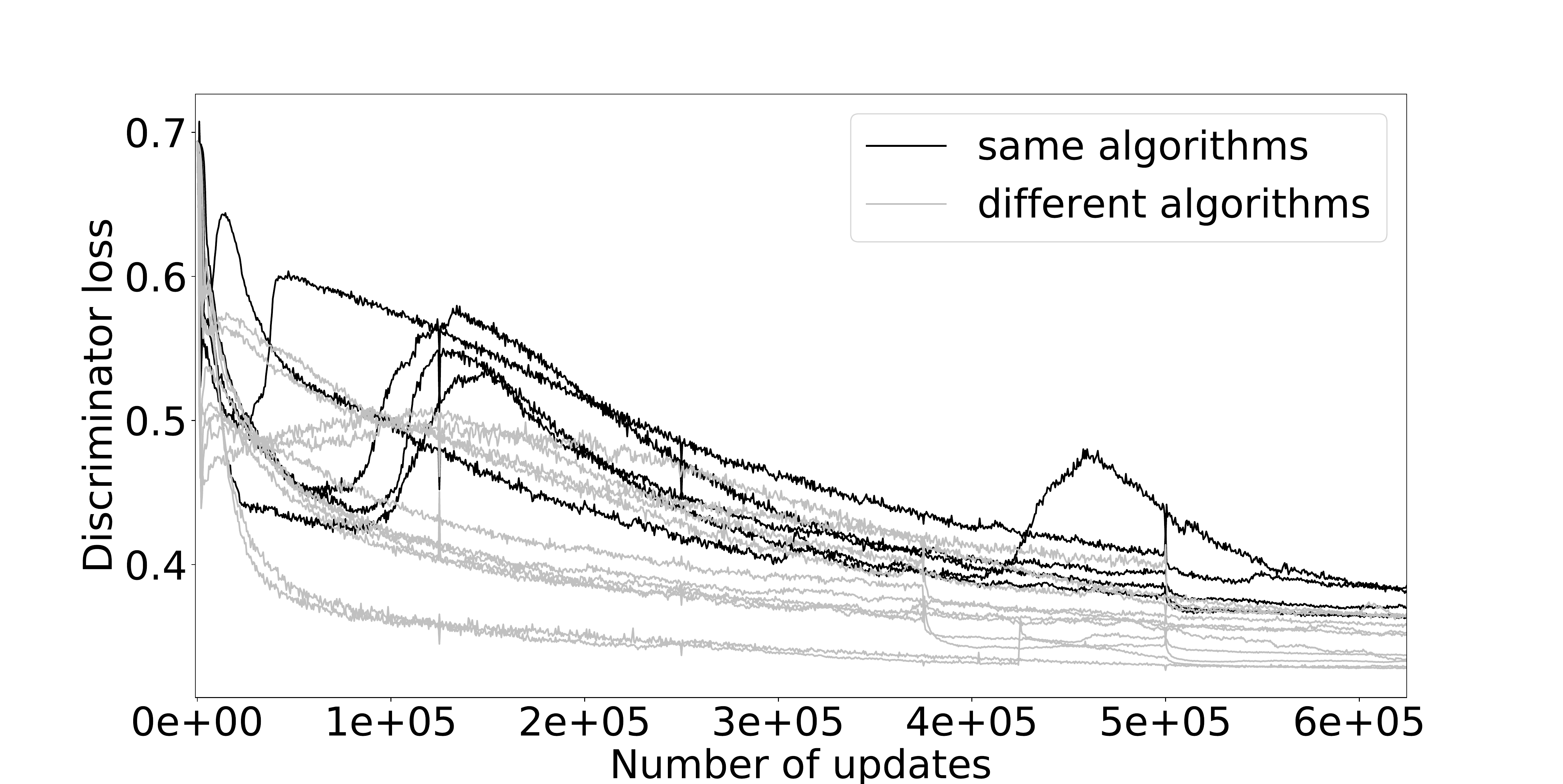}
\caption{Discriminator losses using the same algorithm for source and target (black curves) or using different algorithms (grey curves).}\label{f:losses}
\end{figure}

%\paragraph{Setup}
%Parameters we have to adjust to decreasing amounts of data: number of samples presented to the discriminator.

%\paragraph{Results}
%\begin{itemize}
%\item Alignment quality decreases with data size, so does monolingual embedding quality
%\item Unsupervised and supervised alignment quality are similar
%\end{itemize}

\section{Discussion}

We have shown that the fact that MUSE cannot align two embedding spaces for English induced by different algorithms (even if using the same corpus), is {\em not}~a result of there not being a linear transformation, and not a result of (lack of) normalization or trivial differences in model hyper-parameters. The only explanation left seems to be that the inductive biases of the different algorithms lead to a loss landscape so riddled with local optima that MUSE cannot possible escape them. 

To support this hypothesis, compare the loss curves for the MUSE runs aligning embeddings induced with the {\em same} algorithms (black curves) to the runs aligning embeddings induced with different algorithms, in Figure~\ref{f:losses}. When the embeddings are induced by the same algorithm, we clearly see the contours of a min-max game, suggesting that the generator and discriminator challenge each other, both contributing to a good alignment. When the embeddings are induced by different algorithms, however, the discriminator quickly drops, with the generator unable to push the discriminator out of a local optimum. {\em Understanding when biases induce highly non-convex landscapes, and how to make adversarial training less sensitive to such scenarios, remains an open problem, which we think will be key to the success of unsupervised machine translation and related tasks.}

\bibliography{acl2018}

\begin{thebibliography}{}
\expandafter\ifx\csname natexlab\endcsname\relax\def\natexlab#1{#1}\fi

\bibitem[{Arjovsky and Bottou(2017)}]{Arjovsky:Bottou:17}
Martin Arjovsky and Leon Bottou. 2017.
\newblock Towards principled methods for training generative adversarial
  networks.
\newblock In {\em ICLR\/}.

\bibitem[{Artetxe et~al.(2018)Artetxe, Labaka, and Agirre}]{Artetxe2018}
Mikel Artetxe, Gorka Labaka, and Eneko Agirre. 2018.
\newblock Generalizing and improving bilingual word embedding mappings with a
  multi-step framework of linear transformations.
\newblock In {\em Proceedings of AAAI\/}.

\bibitem[{Bojanowski et~al.(2017)Bojanowski, Grave, Joulin, and
  Mikolov}]{bojanowski2017}
Piotr Bojanowski, Edouard Grave, Armand Joulin, and Tomas Mikolov. 2017.
\newblock Enriching word vectors with subword information.
\newblock {\em Transactions of the Association for Computational Linguistics\/}
  5:135--146.

\bibitem[{Conneau et~al.(2018)Conneau, Lample, Ranzato, Denoyer, and
  J\'{e}gou}]{Conneau:ea:18}
Alexis Conneau, Guillaume Lample, Marc'Aurelio Ranzato, Ludovic Denoyer, and
  Herv\'{e} J\'{e}gou. 2018.
\newblock {Word Translation Without Parallel Data}.
\newblock In {\em Proceedings of ICLR\/}.

\bibitem[{Goodfellow et~al.(2014)Goodfellow, Pouget-Abadie, Mirza, Xu,
  WardeFarley, Ozair, Courville, and Bengio}]{Goodfellow2014}
Ian Goodfellow, Jean Pouget-Abadie, Mehdi Mirza, Bing Xu, David WardeFarley,
  Sherjil Ozair, Aaron Courville, and Yoshua Bengio. 2014.
\newblock Generative adversarial networks.
\newblock In {\em Proceedings of NIPS\/}.

\bibitem[{Levy et~al.(2015)Levy, Goldberg, and Dagan}]{Levy15}
Omer Levy, Yoav Goldberg, and Ido Dagan. 2015.
\newblock Improving distributional similarity with lessons learned from word
  embeddings.
\newblock {\em {TACL}\/} 3:211--225.

\bibitem[{Mikolov et~al.(2013)Mikolov, Chen, Corrado, and Dean}]{Mikolov2013}
Tomas Mikolov, Kai Chen, Greg Corrado, and Jeffrey Dean. 2013.
\newblock {Distributed Representations of Words and Phrases and their
  Compositionality}.
\newblock In {\em Proceedings of NIPS\/}.

\bibitem[{Mikolov et~al.(2013a)Mikolov, Chen, Corrado, and Dean}]{Mikolov2013a}
Tomas Mikolov, Kai Chen, Gregroy~S. Corrado, and Jeffrey Dean. 2013a.
\newblock Efficient estimation of word representations in vector space.
\newblock In {\em Proceedings of ICLR\/}.

\bibitem[{Mimno and Thompson(2017)}]{Mimno:Thompson:17}
David Mimno and Laure Thompson. 2017.
\newblock The strange geometry of skip-gram with negative sampling.
\newblock In {\em Proceedings of EMNLP\/}.

\bibitem[{Mitchell and Steedman(2015)}]{Mitchell:Steedman:15}
Jeff Mitchell and Mark Steedman. 2015.
\newblock Orthogonality of syntax and semantics within distributional spaces.
\newblock In {\em Proceedings of ACL\/}.

\bibitem[{Pennington et~al.(2014)Pennington, Socher, and
  Manning}]{Pennington14}
Jeffrey Pennington, Richard Socher, and Christopher~D. Manning. 2014.
\newblock Glove: Global vectors for word representation.
\newblock In {\em Proceedings of EMNLP\/}.

\bibitem[{Sch\"{o}nemann(1966)}]{Schonemann:66}
Peter Sch\"{o}nemann. 1966.
\newblock A generalized solution of the orthogonal procrustes problem.
\newblock {\em Psychometrika\/} 31:1--10.

\bibitem[{Smith et~al.(2017)Smith, Turban, Hamblin, and Hammerla}]{Smith2017}
Samuel~L. Smith, David H.~P. Turban, Steven Hamblin, and Nils~Y. Hammerla.
  2017.
\newblock {Bilingual word vectors, orthogonal transformations and the inverted
  softmax}.
\newblock In {\em Proceedings of ICLR (Conference Track)\/}.

\bibitem[{S{\o}gaard et~al.(2018)S{\o}gaard, Ruder, and Vulic}]{Soegaard2018}
Anders S{\o}gaard, Sebastian Ruder, and Ivan Vulic. 2018.
\newblock On the limitations of unsupervised bilingual dictionary induction.
\newblock In {\em Proceedings of ACL\/}.

\bibitem[{Zhang et~al.(2017)Zhang, Liu, Luan, and Sun}]{Zhang2017}
Meng Zhang, Yang Liu, Huanbo Luan, and Maosong Sun. 2017.
\newblock Adversarial training for unsupervised bilingual lexicon induction.
\newblock In {\em Proceedings of ACL\/}.

\end{thebibliography}
\bibliographystyle{acl_natbib}

\end{document}